# Lex2vec: making Explainable Word Embeddings via Lexical Resources

**Fabio Celli**

Maggioli Research & Development
via Bornaccino 101
Santarcangelo di Romagna, Italy
`fabio.celli@maggioli.it`

## Abstract

In this technical report, we propose an algorithm, called Lex2vec that exploits lexical resources to inject information into word embeddings and name the embedding dimensions by means of knowledge bases. We evaluate the optimal parameters to extract a number of informative labels that is readable and has a good coverage for the embedding dimensions.

## 1 Introduction and Related Work

From 2000 to 2020, Natural Language Processing adopted several approaches for the study of semantics, from lexical semantics to distributional semantics and word embeddings, context-free or transformer-based. Lexical semantics had the advantage to be fully interpretable, and even the most abstract concepts, like semantic relations [Celli, 2009b] or qualia structures [Pustejovsky and Jezek, 2008] were encoded by labels [Celli, 2010] and classified [Celli, 2009a]. But lexical semantics had great limitations due to the ambiguous nature of words, that required Word-Sense-Disambiguation tasks [Navigli, 2009]. Lexical resources allow us to produce labelled training data for unlabelled sources. Making use of an already existing database, such as Freebase or a domain-specific database, it is possible to collect and label examples for the relation we want to extract. This approach worked very well in semantic relation extraction tasks [Smirnova and Cudré-Mauroux, 2018]. Distributional semantics solved the word ambiguity problem by computing co-occurrence word vectors that made possible to measure the distance between similar words in multidimensional conceptual spaces [Mohammad and Hirst, 2012], opening new possibilities for the extraction of semantic relations [Celli and Nissim, 2009]. Anyway, although the resulting matrices are interpretable, they are also huge and very sparse, and this is a limitation for supervised learning. Context-free Word embeddings [Mikolov *et al.*, 2013] solve the sparsity problem by using neural network representations to embed many word context dimensions into features. Doing so, they reduce the feature space and boost the predictive power in semantic relation extraction tasks [Gábor *et al.*, 2018], but definitely reopen the word disambiguation problem and turn the meaning of each dimension totally opaque. Transformer-based embeddings like BERT [Devlin *et al.*, 2018] perform also word sense disambiaguation because they create a different vector for each different word meaning, but still remain opaque. Crucially, there are efforts towards explainable word embeddings, like EVE, a vector embedding technique which is built upon the structure of Wikipedia and exploits the Wikipedia hierarchy to represent a concept using human-readable labels [Qureshi and Greene, 2019]. Other techniques, like Layerwise Relevance Propagation, try to determine which features in a particular input vector contribute most to a word embedding's output [Şenel *et al.*, 2018], providing some clues for model interpretation but without naming a feature. Existing word embedding learning algorithms typically only use the contexts of words but ignore the sentiment of texts. There are proposals for adding sentiment-specific information into word embeddings [Tang *et al.*, 2015] and this kind of technique could be exploited also for making the embeddings more transparent.

In this technical report, we propose an algorithm, called Lex2vec that exploits lexical resources, like LIWC [Tausczik and Pennebaker, 2010] or NRC [Mohammad *et al.*, 2013] to inject information into word embeddings and name the embedding dimensions by means of distant supervision. In the next section we will present the algorithm, the data and the lexical resources used for the evaluation.

## 2 Algorithm, Data and Evaluation

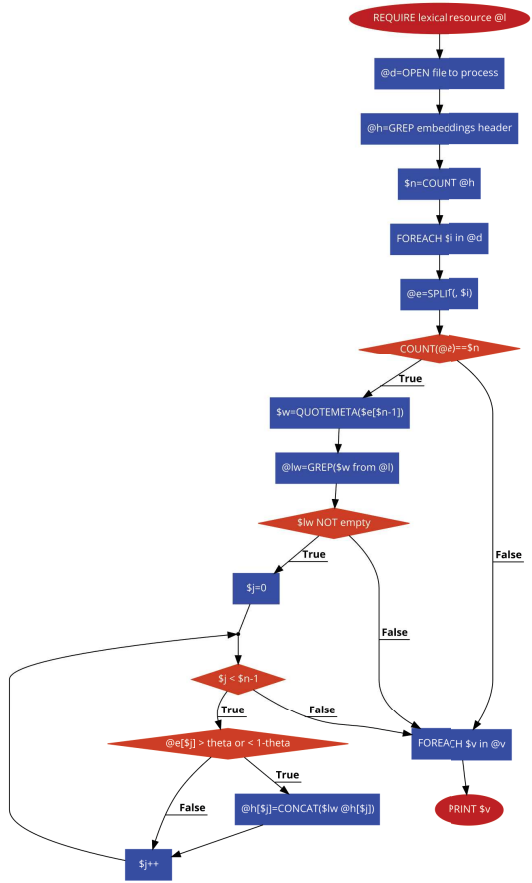

Figure 1: Flowchart of the algorithm Lex2vec.

The Algorithm, depicted as a flowchart in Figure 1, requires a lexical resource `@l` and takes as input a word-embedding dictionary `@d`, produced with Word2Vec or GloVe. All the values in the word-embedding vector must be normalized between 0 and 1. Then the algorithm extracts the header with the unnamed embeddings `@h`, and count how many dimensions there are `$n`. Then for each line `$i` in the dictionary `@d` the algorithm splits the embedding vector `@e`, takes the word `$w` (escaping meta-characters if needed) and check the lexical resource for the corresponding word label(s) `@lw`. If the word label is found, a for loop evaluates if each value of the embedding vector is greater than a threshold `theta` or lower than 1 minus `theta`, and in the case it is, the algorithm maps the label to the corresponding dimension in the header `@h[$j]`, concatenating multiple labels. The threshold `theta` is a parameter that allows us to select the most informative words, the ones that have an embedding score in the highest or in the lowest percentile. The output of the Lex2vec algorithm is a set of labels for each dimension in the header of word embeddings. These labels are mapped from one or more lexical resources with distant supervision, and can be many for some dimensions and none for others. There are many techniques that can filter labels – i.e. a simple limit to the concatenation or a threshold on the ranking of most frequent labels per dimension – but our goal here is to experiment with the theta parameter without filtering techniques, to optimize the number labels (too many labels decrease readability) and reduce the ratio of unnamed dimensions. In order to evaluate the algorithm, we experimented

| theta | resource | % unnamed | avg labels/dim. |
|---|---|---|---|
| 0.81 | liwc | 38.6% | 22.2 |
| 0.79 | liwc | 30.6% | 41.2 |
| 0.77 | liwc | 23.7% | 68.3 |
| 0.75 | liwc | 17.8% | 106.5 |
| 0.81 | nrc | 30.6% | 83.3 |
| 0.79 | nrc | 23.7% | 145.2 |
| 0.77 | nrc | 17.8% | 235.8 |
| 0.75 | nrc | 11.8% | 378.1 |

Table 1: Results of the evaluation.

with ACE2004, a corpus for information extraction from news [Doddington *et al.*, 2004], we extracted word embeddings with Word2vec and applied the Lex2vec algorithm with (a small version of) LIWC (about 500 words) and NRC (about 6400 words) to map words to linguistic labels. Results, reported in

Table 1, show that, as the theta parameter increases, the number of labels per dimension decreases, making the name of the embedding more readable, but the percentage of dimensions that remain unnamed increases as well.

Our conclusion is that the algorithm is suitable for the explainability of word embeddings, and the theta parameter should be around 0.75, suggesting that some strategy to limit the concatenation of labels or select the best ones is necessary, especially with larger lexical resources. This will be material for future work.